%% file: cvpr.tex
\documentclass[final]{cvpr}

\usepackage{times}
\usepackage{epsfig}
\usepackage{graphicx}
\usepackage{amsmath}
\usepackage{amssymb}
\usepackage{bbm}
\usepackage{color}
\usepackage{comment}

\usepackage{xspace}
\newcommand{\LL}{\mathcal{L}}

\newcommand{\indic}[1]{\mathbbm{1}_{[#1]}}

\usepackage[pagebackref=true,breaklinks=true,colorlinks,bookmarks=false]{hyperref}

\usepackage{cleveref}
\crefformat{section}{\S#2#1#3}

\def\cvprPaperID{10021} 
\def\confYear{CVPR 2021}
\begin{document}

\title{Multi-Label Learning from Single Positive Labels}

\author{Elijah Cole$^{1}$\hspace{12pt}Oisin Mac Aodha$^{2}$\hspace{12pt}Titouan Lorieul$^{3}$\hspace{12pt}Pietro Perona$^{1}$\hspace{12pt}Dan Morris$^{4}$\hspace{12pt}Nebojsa Jojic$^{5}$\\
$^{1}$Caltech \hspace{15pt} $^{2}$University of Edinburgh \hspace{15pt} $^{3}$Inria  \hspace{15pt} $^{4}$Microsoft AI for Earth \hspace{15pt}$^{5}$Microsoft Research
}

\maketitle

\begin{abstract}
Predicting all applicable labels for a given image is known as multi-label classification. 
Compared to the standard multi-class case (where each image has only one label), it is considerably more challenging to annotate training data for multi-label classification.
When the number of potential labels is large, human annotators find it difficult to mention all applicable labels for each training image.
Furthermore, in some settings detection is intrinsically difficult \eg finding small object instances in high resolution images. 
As a result, multi-label training data is often plagued by false negatives. 
We consider the hardest version of this problem, where annotators provide only one relevant label for each image.
As a result, training sets will have only one positive label per image and no conﬁrmed negatives. 
We explore this special case of learning from missing labels across four different multi-label image classification datasets for both linear classifiers and end-to-end fine-tuned deep networks. 
We extend existing multi-label losses to this setting and propose novel variants that constrain the number of expected positive labels during training. 
Surprisingly, we show that in some cases it is possible to approach the performance of fully labeled classifiers despite training with significantly fewer confirmed labels.
\end{abstract}

\section{Introduction}

The majority of work in visual classification is focused on the \emph{multi-class} setting, where each image is assumed to belong to one of $L$ classes.
However, the world is intrinsically \emph{multi-label}: scenes contain multiple objects, CT scans reveal multiple health conditions, satellite images show multiple terrain types, etc.
Unfortunately, it can be prohibitively expensive to obtain the large number of accurate multi-label annotations required to train deep neural networks~\cite{deng2014scalable}. 
Heuristics can be used to reduce the required annotation effort~\cite{lin2014coco, gupta2019lvis}, but this runs the risk of increasing error in the labels. 
Even without heuristics, false negatives are common because (i) rare classes are often missed by human annotators~\cite{wolfe2005rare,wolfe2010visual} and (ii) detecting absence can be more difficult than detecting presence~\cite{wolfe2005rare}.
This may explain why even flagship multi-class datasets like ImageNet have been found to include images that actually belong to multiple classes \cite{wu2019tencent}.
Since it is generally infeasible to exhaustively annotate every image for all classes that could be present, there is a natural trade-off between \emph{how many} images receive annotations and \emph{how completely} each image is annotated.
On one extreme, we could fully annotate images until the labeling budget is exhausted.
In this paper we are interested in the other extreme, in which our dataset consists of many images, but each individual image has minimal supervision.

\begin{figure}
    \centering
    \includegraphics[width=0.47\textwidth]{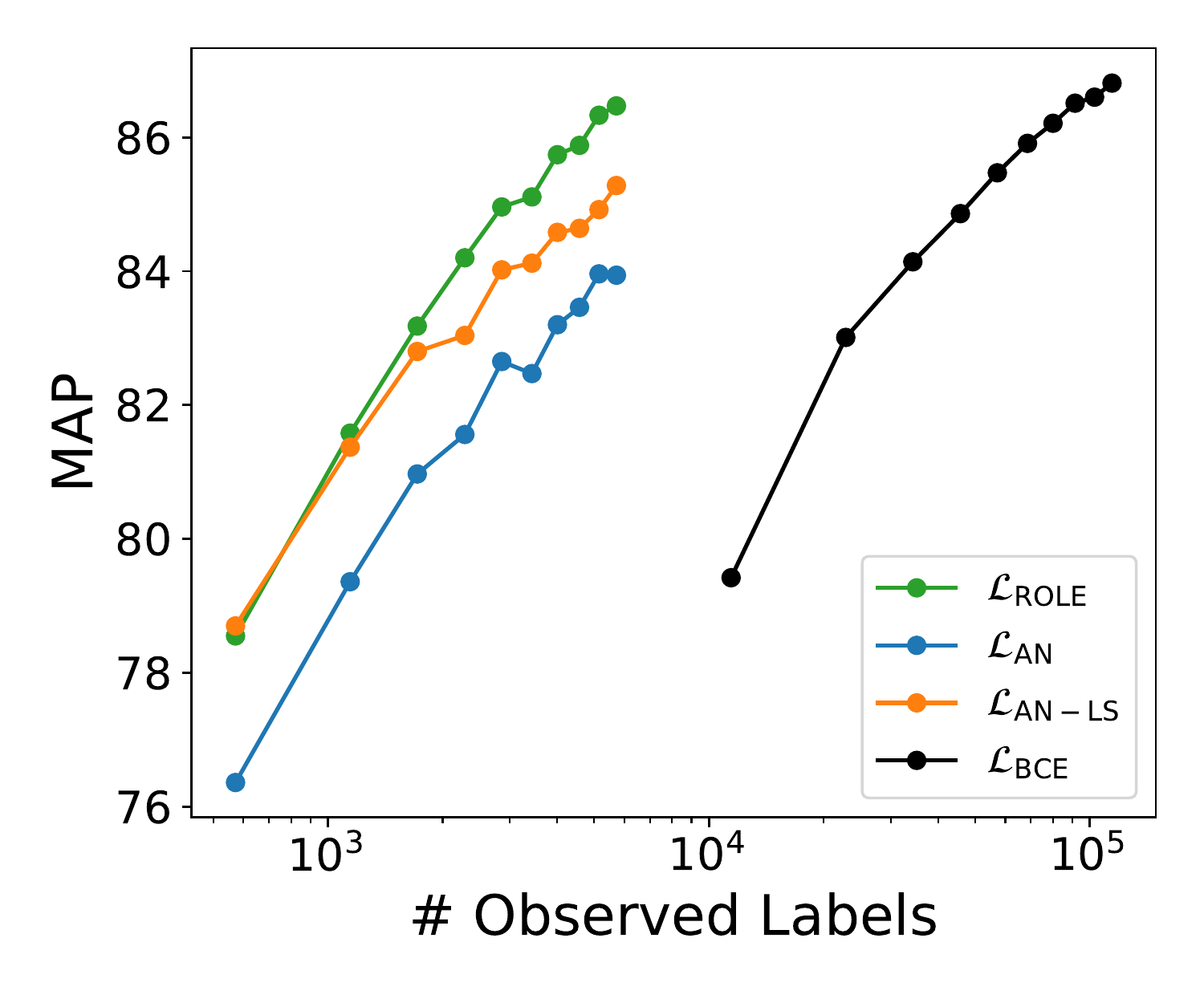}
    \vspace{-10pt}
    \caption{
        It is possible to approach the performance of full supervision ($\mathcal{L}_\mathrm{BCE}$) using only one positive label per image. 
        Here we show test MAP as a function of the number of training labels for PASCAL VOC 2012~\cite{everingham2012pascal}.
        Each curve is generated by randomly subsampling $m\%$ of the images from the training set for $m \in \{10, 20, \ldots, 100\}$.
        The number of labels per image then determines the number of observed label on the horizontal axis: $\mathcal{L}_\mathrm{BCE}$ receives all 20 labels per image, while the other methods only receive one positive label per training image. 
        Despite having a factor of 20 times fewer labels, our $\mathcal{L}_\mathrm{ROLE}$ approach achieves comparable performance to the fully labeled case ($\mathcal{L}_\mathrm{BCE}$).
    }
    \label{fig:compare_obs_labels}
     \vspace{-10pt}
\end{figure}

We explore the problem of {\em single positive multi-label learning}, where only a single positive label (and no other true positives or true negative labels) is observed for each training image. 
This is a worthwhile problem for at least three reasons: 
First, an effective method for this setting could allow for significantly reduced annotations costs for future datasets. 
Second, multi-class datasets may have images that actually contain more than one class.  For instance, the iNaturalist dataset has many images of insects on plants, but only one is annotated as the true class~\cite{van2018inaturalist}. 
Finally, it is of scientific interest to understand how well multi-label classifiers can be made to perform at the minimal limit of supervision.
This is particularly interesting because many standard approaches for dealing with missing labels, \eg learning positive label correlations \cite{chen2019multi}, performing label matrix completion \cite{cabral2011matrix}, or learning to infer missing labels \cite{veit2017learning} break down in the single positive only setting.

We direct attention to this important but underexplored variant of multi-label learning. 
Our experiments show that training with a single positive label per image allows us to drastically reduce the amount of supervision required to train multi-label image classifiers, while only incurring a tolerable drop in classification performance (see Figure~\ref{fig:compare_obs_labels}). 
We make three contributions: 
(i) A unified presentation and extension of existing multi-label approaches to the single positive multi-label learning setting. 
(ii) A novel single positive multi-label loss that estimates missing labels during training.
(iii) A detailed experimental evaluation that compares the performance of multiple different losses across four multi-label image classification datasets. 

\section{Related Work}
Multi-label classification is an important and well studied problem~\cite{zhang2013review,xu2019survey,liu2020emerging} with applications in natural language processing \cite{johnson2015effective,joulin2016bag}, audio classification \cite{briggs2012acoustic,cakir2015polyphonic}, information retrieval~\cite{prabhu2014fastxml}, and computer vision \cite{zhang2007ml,hsu2009multi,gong2014deep,wei2015hcp,wang2016cnnrnn}. 
The conventional approach in vision is to train deep convolution neural networks with multiple output predictions -- one for each concept/class of interest. 
When there are no missing labels (\ie for each image we have complete observations of the presence and absence of each class), standard binary cross-entropy or softmax cross-entropy losses are typically used, \eg~\cite{liu2017deep,mahajan2018exploring}.

In practice, label information is often incomplete at training time because it can be extremely difficult to acquire exhaustive supervision \cite{deng2014scalable}.
Different approaches have been proposed to address the partially labeled setting including: assuming the the missing labels are negative \cite{sun2010multi,bucak2011multi,macaodha2019presence}, ignoring missing labels \cite{durand2019learning}, performing label matrix reconstruction \cite{cabral2011matrix,xu2013speedup}, learning label correlations \cite{chen2019multi,durand2019learning,pineda2019elucidating,huynh2020interactive},  learning generative probabilistic models \cite{kapoor2012multilabel,chu2018deep}, and 
training label cleaning networks \cite{veit2017learning}.
It is worth noting that semi-supervised multi-label classification \cite{liu2006semi,guo2012semi,wang2013dynamic,niu2019multi} can be viewed as a special case of training with missing labels, where here we have entire images with no labels. 
The partially labelled setting is also related to methods that address label noise, \eg \cite{hu2018multilabel,hu2019weakly}. 
Label noise is also encountered in the related area of image tagging \cite{sang2012user, fu2017advances}, where only a small faction of the potentially relevant tags are known for each image.
We are interested in one special kind of label noise,  where some unobserved labels are incorrectly treated as being absent. 
This ``noise'' is the result of a strong assumption, and is not label noise in the traditional sense.
With the exception of the some simple approaches (\eg assuming missing labels are negative \cite{mahajan2018exploring}), most existing approaches assume that they have access to a subset of exhaustively labelled images, or at the very least, images with more than one confirmed positive or negative label. 

We consider a setting where annotators are only asked to provide a single positive label for each training image and no additional negative or positive labels. 
This arises in multi-class image classification where multiple relevant objects may appear in each image but only a single class is annotated \cite{russakovsky2015imagenet}. 
This same problem also occurs in non vision domains such as species distribution modeling~\cite{phillips2004maximum} where the training data are records of real-world (positive) observations for a given location, and there are no negatives. 
The single positive setting has advantages. 
When collecting multi-label annotations, it may be more efficient for a crowd worker to mark the presence of a specific class as opposed to confirming its absence. 

Our setting is most closely related to positive-unlabeled (PU) learning \cite{li2003learning} -- see \cite{bekker2020learning} for a recent survey focused on binary classification, which is the most commonly studied formulation of PU learning. 
In PU learning we only have access to a set of positive items and an additional set of unlabeled items, which may be either positive or negative. 
Compared to the classification setting, there are relatively few works that explore PU learning for multi-label tasks \cite{sun2010multi,hsieh2015pu,kanehira2016multi,han2018multi}, and to the best of our knowledge, there are no works that explicitly explore the single positive case in-depth.
\cite{qiu2017nonconvex} and  \cite{duan2019learning} address the setting where there is only a single label available for each item at training time. 
However, unlike in our setting, these labels can be positive \emph{or} negative. 
Furthermore, when more than one positive label is available for each image, it is possible to infer class level co-occurrence information -- something which is not directly possible with only single positive labels.
In the multi-\emph{class} setting, \cite{ishida2017learning} proposes to learn from complementary labels \ie they assume access to a single negative label per item that specifies that the item does not belong to a given class. 
Their solution falls under the ``assume negative'' set of approaches mentioned earlier, except that the positives and negative labels are reversed. 
Another related multi-class setting is set-valued classification, where each image has one label and the goal is to learn to predict a set of labels as a way to represent uncertainty \cite{chzhen2021set}.
In Section~\ref{po_losses} we discuss several existing multi-label approaches in a unified context and adapt these methods for the single positive setting in Section~\ref{sec:proposed_losses}.

\section{Problem Statement} 
In the standard multi-\emph{class} classification setting, each $\mathbf{x}$ from the input space $\mathcal{X}$ is assigned a single label from $\{1,\ldots,L\}$, where $L$ is the number of classes. 
In the multi-\emph{label} classification setting, each $\mathbf{x}$ is associated with a vector of labels $\mathbf{y}$ from the label space $\mathcal{Y} = \{0, 1\}^L$, where an entry $y_{i} = 1$ if the $i$th class is relevant to $\mathbf{x}$ and $y_{i} = 0$ if the $i$th class is not relevant.

The goal is to find a function $f : \mathcal{X} \to [0, 1]^L$ that predicts the applicable labels for each $\mathbf{x} \in \mathcal{X}$.
The formal objective is to find an $f$ that minimizes the risk 
\begin{align}
    R(f) = \mathbb{E}_{(\mathbf{x},\mathbf{y}) \sim p(\mathbf{x},\mathbf{y})} \overline{\LL} (f(\mathbf{x}),\mathbf{y}),
\end{align} 
where $\overline{\LL} : [0, 1]^L \times \mathcal{Y} \to \mathbb{R}$ reflects some multi-label metric \eg mean average precision or 0-1 error. 
In practice, we define $f$ to be a neural network with parameters $\theta$ and we replace $\overline{\LL}$ with a surrogate $\LL$ that is easier to optimize.
Given an observed dataset $\{(\mathbf{x}_n,\mathbf{y}_n)\}_{n=1}^N$, we can use standard techniques to approximately solve
\begin{align}
    \hat{\theta}_\mathrm{full} = \mathrm{argmin}_{\theta} \frac{1}{N}\sum_{n=1}^N \LL(f(\mathbf{x}_n;\theta),\mathbf{y}_n) \label{eq:erm_true}
\end{align}
where $\LL : [0, 1]^L \times \mathcal{Y} \to \mathbb{R}$ is a suitable multi-label loss function \eg binary cross-entropy or softmax cross-entropy.
However, this formulation assumes that we have access to a \emph{fully observed} label vector $\mathbf{y}_n$ for each input $\mathbf{x}_n$.
In this work we explore the setting where the true label vectors are not directly accessible. 
Instead, during training we observe $\mathbf{z}_n \in \mathcal{Z} = \{0, 1, \varnothing \}^L$, where $z_{ni} \in \{0, 1\}$ is interpreted as before, but $z_{ni} = \varnothing$ indicates that the $i$th label is unobserved for $\mathbf{x}_n$.
That is, if $z_{ni} = \varnothing$ then the corresponding $y_{ni}$ could be either $0$ or $1$.
This is the \emph{partially observed} setting, where we can use our training set $\{(\mathbf{x}_n,\mathbf{z}_n)\}_{n=1}^N$ to approximately solve
\begin{align}
    \hat{\theta}_\mathrm{partial} = \mathrm{argmin}_{\theta} \frac{1}{N}\sum_{n=1}^N \LL(f(\mathbf{x}_n; \theta),\mathbf{z}_n), \label{eq:erm_obs}
\end{align}
where $\LL: [0,1]^L \times \mathcal{Z} \to \mathbb{R}$ is a multi-label loss function that can handle partially observed labels -- see Section \ref{po_losses} for examples. 
Specifically, we focus on a particular instance of the partially observed setting which we call the \emph{single positive only} case, where we observe one single positive label per training example and all the other labels are unknown.
Formally, the single positive case is characterized by 
\begin{align}
    & z_{ni} \in \{1, \varnothing\} \text{ for all } n \in \{1,\ldots,N\}, i \in \{1,\ldots,L\} \nonumber\\
    & \sum_{i=1}^L \indic{z_{ni} = 1} = 1 \text{ for all } n\in\{1,\ldots,N\}
\end{align}
where $\indic{\cdot}$ denotes the indicator function, \ie $\indic{z_{ni} = 1} = 1$ if $z_{ni} = 1$, and $0$ otherwise.
Intuitively we expect a lower risk for the function $f$ learned from fully observed data, \ie $R(f(\cdot; \hat{\theta}_\mathrm{full})) \leq  R(f(\cdot; \hat{\theta}_\mathrm{partial}))$.
The key question is: \emph{how can we design a loss $\LL$ to minimize $R(f(\cdot; \hat{\theta}_\mathrm{partial})) - R(f(\cdot; \hat{\theta}_\mathrm{full}))$?}

\section{Multi-Label Learning}\label{po_losses} 
In this section we compare and contrast three multi-label settings: fully observed labels, partially observed labels (\ie some positives and some negatives are observed), and positive only labels (\ie all observed labels are positive and there are no confirmed negatives).
In the fully observed setting we cover the binary cross-entropy (BCE) loss.
We then discuss how the standard BCE loss is modified to accommodate the partially observed and positive only settings. 
We focus on BCE because it is ubiquitous in multi-label classification, \eg~\cite{veit2017learning,durand2019learning}, but one could carry out a similar exercise using other multi-label losses.
We also compare the different variants in terms of the implicit assumptions each makes regarding unobserved labels.

First we introduce some additional notation. 
Let $\mathbf{f}_n = f(\mathbf{x}_n;\theta) \in [0,1]^L$ be the vector of class probabilities predicted for $\mathbf{x}_n$ by our multi-label classifier $f(\cdot;\theta)$, and let $f_{ni}$ be the $i$th entry of $\mathbf{f}_n$.
Note that since we are using the binary cross-entropy loss, the class probabilities $f_{ni}$ do not sum to one over classes $i$.

\subsection{Fully Observed Labels} 
The binary cross-entropy (BCE) loss is one of the simplest and most commonly used multi-label losses \cite{nam2014largescale,durand2019learning}. 
For a fully observed data point $(\mathbf{x}_n,\mathbf{y}_n)$, the BCE loss is 
\begin{align}
    \LL_\mathrm{BCE}(\mathbf{f}_n, \mathbf{y}_n) = - \frac{1}{L}\sum_{i=1}^L [ &\indic{y_{ni} = 1} \log(f_{ni}) \\
    + & \indic{y_{ni} = 0} \log(1-f_{ni}) ] \nonumber
\end{align}
where we have substituted $\indic{y_{ni} = 1}$ for $P(y_{i} = 1 | \mathbf{x}_n)$ and $\indic{y_{ni} = 0}$ for $P(y_{i} = 0 | \mathbf{x}_n)$. 
In the following sections, we present simple variants of $\mathcal{L}_\mathrm{BCE}$ that do not require fully observed data.
The trade-off is that these variants make stronger implicit assumptions about the distribution $P(y_{i} | \mathbf{x}_n)$.

\subsection{Partially Observed Labels}
Suppose that we have a partially observed data point $(\mathbf{x}_n, \mathbf{z}_n)$.
For observed labels we can simply let $P(y_i = 1 | \mathbf{x}_n) = \indic{z_{ni} = 1}$  and $P(y_i = 0 | \mathbf{x}_n) = \indic{z_{ni} = 0}$ just like we did for $\LL_\mathrm{BCE}$. 
However, it is not clear what to do if a label is unobserved (\ie $z_{ni} = \varnothing$).
One idea is to simply set the loss terms corresponding to unobserved labels to zero, resulting in the ``ignore unobserved" (IU) loss
\begin{align}
    \mathcal{L}_\mathrm{IU}(\mathbf{f}_n, \mathbf{z}_n) = - \frac{1}{L}\sum_{i=1}^L [&\mathbbm{1}_{[z_{ni} = 1]} \log(f_{ni}) \nonumber\\
    + &\mathbbm{1}_{[z_{ni} = 0]} \log(1 - f_{ni})]. \label{eq:assume_perfect}
\end{align}
This loss implicitly assumes that unobserved labels are perfectly predicted, \ie $f_{ni} = P(y_{i} = 1 | \mathbf{x}_n)$ if $z_{ni} = \varnothing$. 
If we additionally weight $\mathcal{L}_\mathrm{IU}(\mathbf{f}_n, \mathbf{z}_n)$ by the number of observed labels in $\mathbf{z}_n$ then we obtain the loss used in \cite{durand2019learning} (up to scaling). 

The $\LL_\mathrm{IU}$ loss allows for missing labels, but it requires both positive and negative labels.
Our focus is the positive-only setting, in which these losses collapse to the trivial ``always predict positive" solution due to the absence of any negative training examples.
Though these losses are inapplicable in our setting, we discuss them to clarify the relationship between our work and \cite{durand2019learning}.
In addition, we use variants of $\mathcal{L}_\mathrm{IU}$ as conceptual tools in our experiments.
In particular, we use a version that ``ignores unobserved negatives" (IUN), given by
\begin{align}
    \mathcal{L}_\mathrm{IUN}(\mathbf{f}_n, \mathbf{z}_n, \mathbf{y}_n) = - \frac{1}{L}\sum_{i=1}^L [&\mathbbm{1}_{[z_{ni} = 1]} \log(f_{ni}) \nonumber\\
    + &\mathbbm{1}_{[y_{ni} = 0]} \log(1 - f_{ni})].
    \label{eq:durand_like}
\end{align}
This is similar to $\LL_\mathrm{IU}$ except with unrealistic access to all of the true negative labels. This hypothetical loss provides an intermediate step between the fully labeled setting and the positive only setting.

\subsection{Positive Only Labels}
Suppose that we have partially observed data $(\mathbf{x}_n, \mathbf{z}_n)$ and suppose that all of the observed labels are positive \ie $z_{ni} \neq \varnothing \implies z_{ni} = 1$. 
We know what to do with observed labels, \ie we set $P(y_i = 1 | \mathbf{x}_n) = \indic{z_{ni} = 1}$.
However, we cannot simply ignore the unobserved labels because that would lead to the degenerate ``always predict positive" solution.
The simplest approach is to assume unobserved labels are negative, \ie $P(y_{ni} = 1 | \mathbf{x}_n) = 0$ if $z_{ni} = \varnothing$. 
The resulting ``assume negative'' (AN) loss is given by 
    \begin{align}
        \mathcal{L}_\mathrm{AN}(\mathbf{f}_n, \mathbf{z}_n) = 
        - \frac{1}{L} \sum_{i=1}^L [&\indic{z_{ni} = 1} \log(f_{ni}) \nonumber \\
        + &\indic{z_{ni} \neq 1} \log(1 - f_{ni}) \label{eq:AN} ].
    \end{align}
This is perhaps the most common approach to the positive only setting, and is explored as ``noisy+" in \cite{durand2019learning}, among others \cite{joulin2015learning, mahajan2018exploring, kundu2020exploiting}.
The drawback is that $\LL_\mathrm{AN}$ will introduce some number of false negatives.
Note that if the role of positive and negative labels are reversed, then this formulation is equivalent to complementary label learning \cite{ishida2017learning}.

\section{Learning From Only Positive Labels}
\label{sec:proposed_losses}

In typical multi-label datasets there are far more negative labels than positive labels.
This means that in the single positive setting, $\mathcal{L}_\mathrm{AN}$ will actually get almost all of the unobserved labels correct.
However, as we demonstrate in our experiments later, even these few false negatives can significantly reduce performance.
An ideal solution to this problem would (i) reduce the damaging effects of false negatives while (ii) retaining as much of the simplicity of $\LL_\mathrm{AN}$ as possible.
With these goals in mind, we propose four ideas for mitigating the impact of false negatives: \emph{weak negatives}, \emph{label smoothing}, \emph{expected positive regularization}, and \emph{online label estimation}.

\subsection{Weak Negatives}
A simple way to reduce the impact of false negatives is to down-weight terms in the loss corresponding to negative labels.
We introduce a weight parameter $\gamma \in [0, 1]$ and define the  ``weak assume negative" (WAN) loss as
    \begin{align*}
        \mathcal{L}_{\mathrm{WAN}}(\mathbf{f}_n, \mathbf{z}_n) = -\frac{1}{L}\sum_{i=1}^L [&\mathbbm{1}_{[z_{ni} = 1]} \log(f_{ni}) \nonumber \\
        + &\mathbbm{1}_{[z_{ni} \neq 1]} \gamma \log(1 - f_{ni}) ].
    \end{align*}
The ``interesting" values of $\gamma$ lie strictly between 0 and 1, since $\gamma = 1$ recovers the standard BCE loss and $\gamma = 0$ admits a trivial solution (``always predict positive").
In the single positive setting, if we choose $\gamma = \frac{1}{L-1}$ then the single positive has the same influence on the loss as the $L-1$ assumed negatives. 
This is similar to the loss used by \cite{macaodha2019presence}, which uses single positive labels to learn spatio-temporal priors for image classification. 
Throughout this paper we use $\gamma = 1 / (L-1)$. 

\textbf{Connection to pseudo-negative sampling.} 
$\LL_{\mathrm{WAN}(\gamma)}$ has a probabilistic interpretation based on sampling negatives at random.
Consider the following procedure: each time $(\mathbf{x}_n, \mathbf{z}_n)$ occurs in a batch, choose one of the $L-1$ unobserved labels uniformly at random and treat it as negative.
We repeat this step each time the pair $(\mathbf{x}_n, \mathbf{z}_n)$ appears in a batch.
Since there are typically many more negatives than positives for a given image, our randomly chosen \emph{pseudo-negative} will be a true negative more often than not.
Since we now have both positive and negative labels, we can use the $\mathcal{L}_\mathrm{IU}$ loss, resulting in
    \begin{align*}
        -\frac{1}{L}\sum_{i=1}^L [\mathbbm{1}_{[z_{ni} = 1]} \log(f_{ni}) + \mathbbm{1}_{[z_{ni} \neq 1]} \eta_{ni} \log(1 - f_{ni}) ]
    \end{align*}
where $\eta_{ni}$ is a random variable which is $1$ if $z_{ni}$ is chosen as the pseudo-negative and $0$ otherwise. 
If we take the expectation with respect to the pseudo-negative sampling then we recover $\mathcal{L}_\mathrm{WAN}$ with $\gamma = \frac{1}{L-1}$. 
Though the two losses are equivalent in expectation, they may differ significantly in practice. 

\subsection{Label Smoothing} 
Label smoothing was proposed in \cite{szegedy2016rethinking} as a way to reduce overfitting when training multi-class classifiers with the categorical cross-entropy loss. 
Label smoothing has since been shown to mitigate the effects of label noise in the multi-class setting \cite{muller2019when}.
If we reframe $\mathcal{L}_\mathrm{AN}$ as $\mathcal{L}_\mathrm{BCE}$ with some  ``noisy'' labels (\ie those labels incorrectly assumed to be negative), then it is natural to ask whether label smoothing could help to reduce the impact of those incorrect labels.

In a multi-class context, the target distribution $\textbf{y}_n$ is a delta distribution supported on the correct class label.
Label smoothing replaces $\textbf{y}_n$ with $(1-\epsilon) \textbf{y}_n + \epsilon \textbf{u}$ where $\textbf{u} = [1/L, \ldots, 1/L]$ is the discrete uniform distribution with support size $L$ and $\epsilon \in (0, 1)$ is a hyperparameter.
It is possible to generalize traditional multi-class label smoothing to the binary cross-entropy loss, by simply applying label smoothing independently to each of the $L$ binary target distributions $(\indic{z_{ni} \neq 1}, \indic{z_{ni}, = 1})$. 
We refer to the combination of the ``assume negative'' loss from Eqn.~\ref{eq:AN} with label smoothing as
    \begin{align}
        \mathcal{L}_{\mathrm{AN-LS}}(\mathbf{f}_n, \mathbf{z}_n) = - \frac{1}{L}&\sum_{i=1}^L [ \indic{z_{ni}=1}^{\frac{\epsilon}{2}} \log(f_{ni}) \nonumber \\
        + &\indic{z_{ni} \neq 1}^{\frac{\epsilon}{2}} \log(1-f_{ni})  ],
    \end{align}
where $\epsilon$ is the label smoothing parameter and $\indic{Q}^{\alpha} = (1-\alpha)\indic{Q} + \alpha \indic{\neg Q}$ for any logical proposition $Q$. Throughout this paper we use $\epsilon = 0.1$.

\subsection{Expected Positive Regularization}\label{exp_pos_reg}
Another way to avoid the label noise introduced by assuming unobserved labels are negative is to apply a loss to only the observed labels as in \cite{durand2019learning}. 
However, in the positive only case the loss would be
\begin{align*}
    \mathcal{L}_\mathrm{BCE}^+(\mathbf{f}_n, \mathbf{z}_n) = -\sum_{i=1}^L \indic{z_{ni} = 1} \log(f_{ni}),
\end{align*}
which has a trivial solution, \ie predict that every label is positive.
We propose to build some domain knowledge into the loss to avoid this problem.
Let us assume we have access to a scalar $k$, which is defined as the expected number of positive labels per image:
\[k = \mathbb{E}_{(\mathbf{x},\mathbf{y}) \sim p_\mathrm{data}(\mathbf{x},\mathbf{y})}\sum_{i=1}^L \mathbbm{1}_{[y_i = 1]}.\] 
We can estimate $k$ from data or treat it as a hyperparameter.

Suppose we draw a batch of images with indices $B \subset \{1,\ldots, N\}$.
We define $\mathbf{F}_B = [f_{ni}]_{n\in B, i\in\{1,\ldots,L\}}$ to be the matrix of predictions $f_{ni} \in [0,1]$ for every image in the batch and category in the dataset.
We can use the batch predictions $\mathbf{F}_B$ to compute
\[ \hat{k}(\mathbf{F}_B) = \frac{\sum_{n\in B} \sum_{i=1}^L \mathbf{f}_{ni}}{|B|}.\]
Ideally we would make perfect predictions, \ie $\mathbf{F}_B = \mathbf{Y}_B$ where $\mathbf{Y}_B = [y_{ni}]_{n\in B, i\in\{1,\ldots,L\}}$ is the matrix of true labels.
A necessary condition for $\mathbf{F}_B = \mathbf{Y}_B$ is $\mathbb{E}[\hat{k}(\mathbf{F}_B)] = \mathbb{E}[\hat{k}(\mathbf{Y}_B)]$, where the expectation is taken over batch sampling.
Since $\mathbb{E}[\hat{k}(\mathbf{Y}_B)] = k$ by the definition of $k$, it makes sense to introduce a regularization term $R_k(\mathbf{F}_B)$ that encourages $\hat{k}(\mathbf{F}_B)$ to be close to $k$. We can use this regularizer to implicitly penalize negatives and avoid the trivial ``always predict positive" solution, leading to the loss
\begin{align}
    \mathcal{L}_\mathrm{EPR}(\mathbf{F}_B, \mathbf{Z}_B) &= \frac{1}{|B|}\sum_{n\in B} \mathcal{L}_\mathrm{BCE}^+(\mathbf{f}_n, \mathbf{z}_n) + \lambda R_k(\mathbf{F}_B), \nonumber
\end{align}
where $\lambda$ is a hyperparameter. 
Regularizing at the batch level (instead of the image level) respects the fact that some images will have more than $k$ positive labels and some will have fewer. 

How should we define $R_k(\mathbf{F}_B)$?
Since the number of classes $L$ can vary widely depending on the dataset, we propose to work with the normalized deviation $(\hat{k}(\mathbf{F}_B) - k) / L \in [-1, 1]$.
Penalizing this relative deviation makes sense in contexts where \eg an absolute deviation of 1 matters more if $L=10$ than it does if $L=100$.
We can then define a variety of regularizers with any standard functional form.
We use the squared error, leading to
\begin{align}
    R_k(\mathbf{F}_B) &= \left(\frac{\hat{k}(\mathbf{F}_B) - k}{L}\right)^2.
\end{align}

\subsection{Online Estimation of Unobserved Labels}

While the idea behind $\mathcal{L}_\mathrm{EPR}$ seems reasonable, we find that it does not work well in our experiments (see Section~\ref{sec:experiments}).
In this section we combine $\mathcal{L}_\mathrm{EPR}$ with a second module which maintains online estimates of the unobserved labels throughout training. 
The resulting method is similar to an expectation-maximization algorithm which jointly trains the image classifier and estimates the labels subject to constraints imposed by $\mathcal{L}_\mathrm{EPR}$. 
We refer to this technique as \emph{regularized online label estimation} (ROLE).

To make this more precise we will need some additional notation. We write the estimated labels as $\tilde{\mathbf{Y}} \in[0,1]^{N\times L}$ in analogy with the matrix of true labels $\mathbf{Y}\in\{0,1\}^{N \times L}$ and the matrix of classifier predictions $\mathbf{F} \in [0, 1]^{N \times L}$.
We carry through the derived notation: $\tilde{\mathbf{Y}}_B \in [0,1]^{|B| \times L}$ for a batch $B$, $\tilde{\mathbf{y}}_n \in [0,1]^L$ for a row, and $\tilde{y}_{ni}\in[0,1]$ for a single entry.
Finally, we make the (non-restrictive) assumption that $\tilde{\textbf{y}}_{n} = g(\mathbf{x}_n; \phi)$ where the \emph{label estimator} $g:~\mathcal{X} \to [0,1]^L$ is some function with parameters $\phi$.
We discuss our implementation of $g$ later.

With this notation, our goal is to jointly train the label estimator $g(\cdot; \phi)$ and the image classifier $f(\cdot; \theta)$.
We first consider the intermediate loss
\begin{align}
    \mathcal{L}'(\mathbf{F}_B | \tilde{\mathbf{Y}}_B) &= \frac{1}{|B|} \sum_{n\in B} \mathcal{L}_\mathrm{BCE}(\mathbf{f}_n, \mathrm{sg}(\tilde{\mathbf{y}}_n)) \nonumber \\
    &+ \mathcal{L}_\mathrm{EPR}(\mathbf{F}_B, \mathbf{Z}_B),\label{eq:spo_prime}
\end{align}
where $\mathrm{sg}$ is the stop-gradient function which prevents its argument from backpropagating gradients~\cite{grill2020bootstrap} and we have suppressed the dependence on $\mathbf{Z}_B$ on the left-hand side because $\mathbf{Z}$ is fixed throughout training.
The $\mathcal{L}_\mathrm{BCE}$ term encourages the image classifier predictions $\mathbf{F}_B$ to match the estimated labels $\tilde{\mathbf{Y}}_B$, while the $\mathcal{L}_\mathrm{EPR}$ term pushes $\mathbf{F}_B$ to correctly predict known positives and respect the expected number of positives per image.
We can use this loss to update $\theta$ while assuming that $\phi$ is fixed.
By switching the arguments in Eqn.~\ref{eq:spo_prime} we obtain an analogous loss which allows us to update $\phi$ while assuming $\theta$ is fixed.
Then our final loss is simply 
\begin{align*}
    \mathcal{L}_\mathrm{ROLE}(\mathbf{F}_B, \tilde{\mathbf{Y}}_B) &=
    \frac{\mathcal{L}'(\mathbf{F}_B | \tilde{\mathbf{Y}}_B) + \mathcal{L}'(\tilde{\mathbf{Y}}_B | \mathbf{F}_B)}{2}
\end{align*}
through which we can update $\mathbf{F}_B$ and $\tilde{\mathbf{Y}}_B$ simultaneously.

We now give some intuition for why this might work.
We start with an informal proposition: all else being equal, a convolutional network will more readily train on informative labels than on uninformative labels. 
Concretely, it has been observed that convolutional neural networks can be trained to accurately predict completely random labels, but the same network will fit to the correct labels much faster~\cite{understanding2017zhang}. 
How does this relate to our context?
$\mathcal{L}_\mathrm{ROLE}$ allows the labels to be set arbitrarily, as long as they are consistent with the known labels and the expected number of positive labels.
Since it is easier to train image classifiers on informative labels than uninformative ones, we hypothesize that \emph{correct labels are a ``good choice" from the algorithm's perspective}.
While it is possible to learn to predict labels unrelated to the image content, in many cases it may be easier to predict the correct ones. 

\section{Experiments}\label{sec:experiments} 
Here we present multi-label image classification results on four standard benchmark datasets: PASCAL VOC 2012 (VOC12) \cite{everingham2012pascal}, MS-COCO 2014 (COCO) \cite{lin2014coco}, NUS-WIDE (NUS) \cite{chua2009nus}, and CUB-200-2011 (CUB) \cite{WahCUB_200_2011}. 
For each dataset we present results for both (i) linear classification on fixed features and (ii) end-to-end fine-tuning.

\begin{figure}[t]
    \centering
    \includegraphics[trim={25pt 0 0pt 25pt},clip,width=0.50\textwidth]{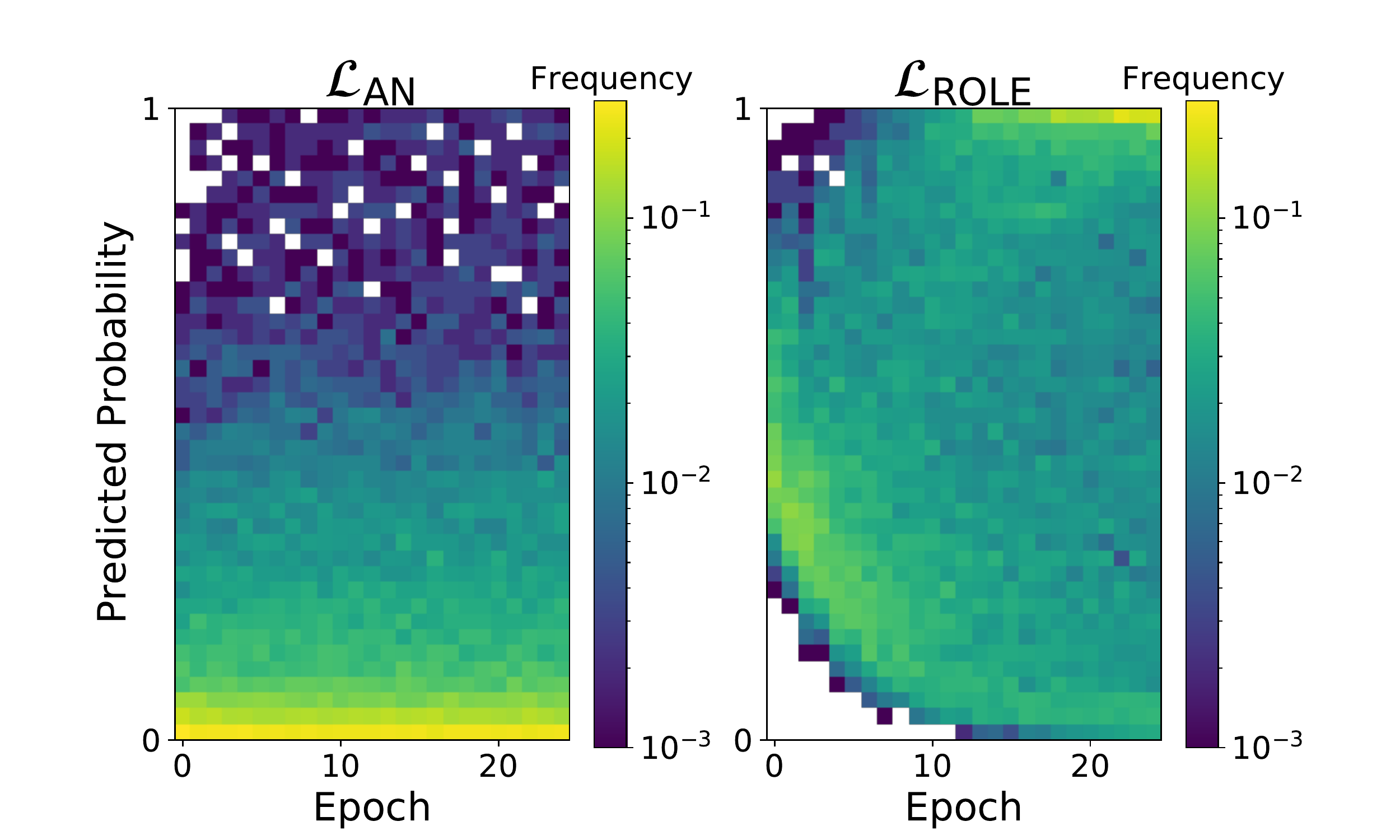}
    \vspace{-10pt}
    \caption{
        Distribution of predicted probabilities for \emph{unobserved} positives when training with a single positive per image for COCO. Each column represents a normalized histogram and white pixels indicate a frequency of zero. 
        Training with $\mathcal{L}_\mathrm{ROLE}$ (right) results in the recovery of a significant number of the unlabeled positives as evident by the majority of the probability correctly being concentrated at 1.0 (top right) by the end of training. 
        $\LL_\mathrm{AN}$ (left) does not exhibit the same behavior.
    }
    \label{fig:pos_distribution}
\end{figure}

\begin{table*}[t]
\centering
\small
\begin{tabular}{|l|l|c c c c| c c c c |}
\hline
    \multicolumn{2}{|c}{} & \multicolumn{4}{|c|}{Linear} & \multicolumn{4}{c|}{Fine-Tuned} \\ \hline
     Loss & Labels Per Image & VOC12 & COCO & NUS & CUB & VOC12 & COCO & NUS & CUB\\ \hline
     $\mathcal{L}_\mathrm{BCE}$ & All Pos. \& All Neg. & 86.7 & 70.0 & 50.7 & 29.1 & 89.1 & 75.8 & 52.6 & 32.1 \\ 
     $\mathcal{L}_\mathrm{BCE-LS}$ & All Pos. \& All Neg. & 87.6 & 70.2 & 51.7 & 29.3 & 90.0 & 76.8 & 53.5 & 32.6 \\
     $\mathcal{L}_\mathrm{IUN}$ & 1 Pos. \& All Neg. & 86.4 & 67.0 & 49.0 & 19.4 & 87.1 & 70.5 & 46.9 & 21.3 \\ 
     $\mathcal{L}_\mathrm{IU}$ & 1 Pos. \& 1 Neg. & 82.6 & 60.8 & 43.6 & 16.1 & 83.2 & 59.7 & 42.9 & 17.9 \\ \hline 
     $\mathcal{L}_\mathrm{AN}$ & 1 Pos. \& 0 Neg. & 84.2 & 62.3 & 46.2 & 17.2 & 85.1 & 64.1 & 42.0 & 19.1 \\
     $\mathcal{L}_\mathrm{AN-LS}$ & 1 Pos. \& 0 Neg. & \underline{85.3} & \underline{64.8} & \underline{48.5} & 15.4 & 86.7 & 66.9 & 44.9 & 17.9 \\
     $\mathcal{L}_{\mathrm{WAN}}$ & 1 Pos. \& 0 Neg. & 84.1 & 63.1 & 45.8 & \underline{17.9} & 86.5 & 64.8 & 46.3 & \textbf{20.3}\\
     $\mathcal{L}_\mathrm{EPR}$ & 1 Pos. \& 0 Neg. & 83.8 & 62.6 & 46.4 & \textbf{18.0} & 85.5 & 63.3 &  46.0 & \underline{20.0} \\
     $\mathcal{L}_\mathrm{ROLE}$ & 1 Pos. \& 0 Neg. & \textbf{86.5} & \textbf{66.3} & \textbf{49.5} & 16.2 & \underline{87.9} & 66.3 & 43.1 & 15.0\\ 
     $\mathcal{L}_\mathrm{AN-LS}$ $+ \mathrm{Linear  Init.}$ & 1 Pos. \& 0 Neg. & - & - & - & - & 86.5 & \textbf{69.2} & \underline{50.5} & 16.6 \\
     $\mathcal{L}_\mathrm{ROLE}$ \hspace{3.0pt} $+ \mathrm{Linear  Init.}$ & 1 Pos. \& 0 Neg. & - & - & - & - & \textbf{88.2} & \underline{69.0} & \textbf{51.0} & 16.8 \\
      \hline
\end{tabular}
\vspace{5pt}
\caption{
Multi-label test set mean average precision (MAP) for different multi-label losses on four different image classification datasets.
We present results for two scenarios: (i) training a linear classifier on fixed features and (ii) fine-tuning the entire network end-to-end. 
In all cases the backbone network is an ImageNet pre-trained ResNet-50. 
All methods below the break use only one positive per image (\ie 1 Pos. \& 0 Neg.), while methods above the break use additional supervision.
In each column we bold the best performing single positive method and underline the second-best.
For each method and we select the hyperparameters that perform the best on the held-out validation set. 
For losses labeled with ``$\mathrm{LinearInit.}$" we freeze the weights of the backbone network for the initial epochs of training and then fine-tune the entire network end-to-end for the remaining epochs.
Note that this linear initialization phase is identical to the training protocol for the ``Linear" results.
}
\label{tab:natimg_compare}
\end{table*}

\begin{table}[t]
\centering
\small
\begin{tabular}{| l | c c c c|} \hline
     Loss & VOC12 & COCO & NUS & CUB\\ \hline
     $\mathcal{L}_\mathrm{AN}$ & 85.8 & 63.8 & 49.3 & 16.8 \\
     $\mathcal{L}_\mathrm{AN-LS}$ & 86.9 & 65.4 & 49.7 & 17.4 \\
     $\mathcal{L}_\mathrm{ROLE}$ & \textbf{90.3} & \textbf{69.5} & \textbf{56.0} & \textbf{19.6}  \\ \hline
\end{tabular} 
\vspace{5pt}
\caption{
Training set MAP for multi-label predictions evaluated with respect to the \emph{full} ground truth labels. 
These values measure how well each method recovers the true training labels despite being trained with one positive label per image. 
Note that all results are for the linear case.
Hyperparameters and stopping epoch are selected using the validation set as before. 
}
\label{tab:label_recovery}
\end{table}

\subsection{Implementation Details}

\textbf{Data preparation.} 
Our goal is to evaluate the performance of different single positive multi-label learning losses. 
To do this, we begin with fully labeled multi-label image datasets and corrupt them by discarding annotations. 
Specifically, we simulate single positive training data by randomly selecting one positive label to keep for each training example. 
This is performed once for each dataset and the same label set is used for all comparisons on that dataset, \ie  every time an image appears in a batch it has the same single positive label.
For each dataset, we withhold 20\% of the training set for validation.
The validation and test sets are always fully labeled.
VOC12 contains 5,717 training images and 20 classes, and we report results on the official validation set (5,823 images). 
COCO consists of 82,081 training images and 80 classes, and we also report results on the official validation set (40,137 images). 
The complete NUS dataset is not available online so we re-scraped it from Flickr. 
As a result, it was not possible to obtain all of the original images. 
In total, we collected 126,034 and 84,226 images from the official training and test sets respectively, consisting of 81 classes.
In accordance with standard practice \cite{gong2014deep,durand2019learning}, we merged the training and test sets and randomly selected 150,000 images for training and used the remaining 60,260 for testing.
CUB is divided into 5,994 training images and 5,794 test images. 
Each CUB image is associated with a vector indicating the presence or absence of 312 binary attributes.
Note that subsets of these attributes are known to be mutually exclusive, but we do not make use of that information.
We provide additional statistics on the datasets in the supplementary material. 

\textbf{Hyperparameters.}
For each method, we conducted a hyperparameter search and selected the hyperparameters with the best mean average precision (MAP) on the validation set. 
We considered learning rates in $\{1e-2, 1e-3, 1e-4, 1e-5\}$ and batch sizes in $\{8, 16\}$.
We train for 25 epochs in the linear case and 10 epochs in the fine-tuned case.
The rows tagged with ``+LinearInit" are fine-tuned for 5 epochs starting from the best weights found during linear training. 
For $\mathcal{L}_\mathrm{ROLE}$ we set the learning rate for the label estimate parameters $\phi$ to be $10\times$ larger than the learning rate for the image classifier parameters $\theta$.
For $\mathcal{L}_\mathrm{EPR}$ and $\mathcal{L}_\mathrm{ROLE}$ we compute $k$ based on the fully labeled training set - we give these values and study the effect of mis-specifying $k$ in the supplementary material.
All experiments are based on a ResNet-50~\cite{he2016deep} pre-trained on ImageNet~\cite{russakovsky2015imagenet}. 

\textbf{Implementation of $g$ for $\mathcal{L}_\mathrm{ROLE}$.} 
We let $\phi \in [0,1]^{N\times L}$ and define $\tilde{\mathbf{Y}}$ by $\tilde{y}_{ni} = \sigma(\phi_{ni})$ where $\sigma: \mathbb{R} \to (0, 1)$ is the sigmoid function.
As a result, $g$ is a simple ``look-up" operation given by $g(\mathbf{x}_n; \phi) = \tilde{\mathbf{y}}_n$. 
We initialize $\phi_{ni}$ from the uniform distribution on $[\sigma^{-1}(0.4), \sigma^{-1}(0.6)]$ if $z_{ni} = 0$ or we initialize $\phi_{ni} = \sigma^{-1}(0.995)$ if $z_{ni} = 1$.
Note that this does not apply to ``$\mathcal{L}_\mathrm{ROLE}$+LinearInit." which starts from the $\phi$ parameters found during linear training.

\subsection{Single Positive Classification Results} 
In Table \ref{tab:natimg_compare} we evaluate the different training losses outlined earlier in the paper in the single positive case (\ie ``1 Pos. \& 0 Neg.'') and compare their performance to other labeling regimes (\eg fully labeled, ``All Pos. \& All Neg.''). 
We also compare against intermediate variants such as  $\mathcal{L}_\mathrm{IUN}$, which has access to one positive label per image and all the negatives \ie more labels than the single positive case, but fewer than the fully labeled case.
We find that $\mathcal{L}_\mathrm{ROLE}$ is the strongest method in the linear case, often approaching (and sometimes surpassing) the performance of $\mathcal{L}_\mathrm{IUN}$, which has access to many more labels at training time.
In the fine-tuned case, we see that better initialization provides substantial benefits to both $\mathcal{L}_\mathrm{ROLE}$ and $\mathcal{L}_\mathrm{AN}$ (see rows with ``+LinearInit.").
However, $\mathcal{L}_\mathrm{AN-LS}$ is also very effective, especially in light of its simplicity. 

\textbf{Single positive training performs surprisingly well.}
One way to better understand the overall performance is by comparing different losses in terms of the number of training labels used. 
In Figure~\ref{fig:compare_obs_labels} we observe that in the linear case, $\mathcal{L}_\mathrm{ROLE}$ achieves test MAP comparable to the fully labelled loss ($\mathcal{L}_\mathrm{BCE}$) on VOC12, despite using 20 times fewer labels.

\textbf{The choice of single positive loss matters.}
While we have discussed the shortcomings of the assume negative baseline  $\mathcal{L}_\mathrm{AN}$, we observe that it performs reasonably well. 
However, we note that the gap between $\mathcal{L}_\mathrm{AN}$ and the fully supervised $\mathcal{L}_\mathrm{BCE}$ is substantially wider in the end-to-end fine-tuned case.
Presumably this is due to the fact that the false negative labels can do much more damage when they are able to corrupt the backbone feature extractor.
This result adds to a broader conversation (which has mostly been focused on the multi-class setting) about whether, and to what extent, deep learning is robust to label noise \cite{rolnick2017deep}.
Our multi-label label smoothing variant $\mathcal{L}_\mathrm{AN-LS}$ and our $\mathcal{L}_\mathrm{ROLE}$ loss perform much better in most cases, indicating that the widely used $\mathcal{L}_\mathrm{AN}$ baseline is a lower bound on performance. 
We also note that although $\mathcal{L}_\mathrm{EPR}$ typically performs worse than $\mathcal{L}_\mathrm{AN}$, it seems to work quite well for CUB.
CUB is unusual among our datasets because the average number of positive labels per image is over 30 (more than $10\times$ higher than VOC12, COCO, and NUS).
We suspect that the relatively mild loss applied to unobserved labels under $\mathcal{L}_\mathrm{EPR}$ may be beneficial when there are so many unobserved positives.

\textbf{Label smoothing is a strong baseline.}
\cite{lukasik2020does} showed that label smoothing mitigates the damaging effects of label noise in the multi-class setting.
We extend these results to the multi-label setting.
We see in Table~\ref{tab:natimg_compare} that $\mathcal{L}_{\mathrm{AN-LS}}$ (\ie assume negative with label smoothing) outperforms the basic assume negative $\mathcal{L}_{\mathrm{AN}}$ loss in nearly every case.
It is also worth noting that label smoothing provides a larger benefit in the single positive case ($\mathcal{L}_{\mathrm{AN-LS}}$ vs. $\mathcal{L}_{\mathrm{AN}}$)  than it does in the fully labeled case ($\mathcal{L}_{\mathrm{BCE-LS}}$ vs. $\mathcal{L}_{\mathrm{BCE}}$).
We therefore recommend $\mathcal{L}_\mathrm{AN-LS}$ as a strong and simple baseline for the single positive multi-label setting. 
However, training with our $\mathcal{L}_{\mathrm{ROLE}}$ loss still performs best in most settings. 
$\mathcal{L}_{\mathrm{ROLE}}$ requires more parameters to be estimated at training time, but incurs no additional computational overhead at inference time. 
In Table~\ref{tab:label_recovery} we present MAP scores computed on the fully observed training set for losses trained with only a single positive per image. 
Interestingly, we observe that $\mathcal{L}_{\mathrm{ROLE}}$ does a better job at recovering the full unobserved label matrix when compared to $\mathcal{L}_{\mathrm{AN-LS}}$. 
This is illustrated qualitatively in Figure~\ref{fig:pos_distribution}, which shows that $\LL_\mathrm{ROLE}$ can successfully recover many of the unobserved positive labels during training. 
However, as seen in Table~\ref{tab:natimg_compare}, this better recovery of the clean training labels does not necessarily translate to comparable gains on the test set.  

\textbf{Initialization matters.}
$\mathcal{L}_{\mathrm{ROLE}}$ is very effective in the linear setting \ie when training a randomly initialized linear classifier and label estimator on frozen backbone features.
However, we find that starting from a randomly initialized classifier and label estimator in the end-to-end setting results in an inferior model. 
This is perhaps not too surprising given the additional degrees of freedom afforded by end-to-end fine-tuning. 
However, as a simple remedy we recommend starting with a frozen backbone for the first few epochs of end-to-end training, which is denoted in Table~\ref{tab:natimg_compare} as $\mathcal{L}_\mathrm{ROLE} + \mathrm{Linear  Init}$. 
We observe that this procedure also provides substantial benefits for $\mathcal{L}_{\mathrm{AN-LS}}$, the label smoothed version of the assume negative training loss. 

\section{Limitations}
When creating our simulated training annotations, our single positive label generation process assumes that for a given image any positive label that is present is equally likely to be annotated. 
This is in line with similar assumptions made in other related work \eg ~\cite{durand2019learning}. 
However, in practice this is an oversimplification, as human annotators are likely to have biases related to the object categories they annotate. 
Depending on the specific dataset, this could be manifested as a preference for annotating familiar object categories, or it could be based on factors related to the saliency of the object instance in the image \eg smaller objects may be less likely to be annotated compared to larger ones. 
In this work we focus on better understanding the potential of single positive training, and leave modeling annotation biases to future work. 

Our $\mathcal{L}_\mathrm{ROLE}$ loss requires the online estimation of an $N\times L$ label matrix. 
As presented, we store the full label matrix in memory.
For a dataset like ImageNet this would require 4GB of memory, but would become infeasible for larger datasets or larger numbers of labels. 
Possible alternative implementations of the label estimator $g$ (which would still be fully compatible with our loss) include learning a factorized estimate of the matrix or using a small neural network to approximate it.  

\section{Conclusion}
We have investigated an underexplored variant of partially observed multi-label classification -- that of single positive training.
Perhaps surprisingly, we have showed that in this supervision deprived setting it is possible to achieve classification results that are competitive with full label supervision using an order of magnitude fewer labels. 
This opens up future avenues of work related to efficient crowdsourcing of annotations for large-scale multi-label datasets. 
In future work we intend to further explore the connections to semi-supervised multi-label classification along with  applications in self-supervised representation learning where the problem of how to address false negative labels often occurs.
In addition, many of the ideas  discussed are applicable to the more general ``partially observed multi-label" case (\ie not just positive labels), and we plan to consider extensions to that setting also.

\vspace{5pt}
\noindent{\bf Acknowledgements.}
This project was supported in part by an NSF Graduate Research Fellowship (Grant No. DGE1745301) and the Microsoft AI for Earth program. We would also like to thank Jennifer J. Sun, Matteo Ruggero Ronchi, and Joseph Marino for helpful feedback.

{\small
\bibliographystyle{ieee_fullname}
\bibliography{cvpr}
}

\clearpage
\appendix
\setcounter{table}{0}
\renewcommand{\thetable}{A\arabic{table}}
\setcounter{figure}{0}
\renewcommand{\thefigure}{A\arabic{figure}}
\setcounter{equation}{0}
\renewcommand{\theequation}{A\arabic{equation}}
\input{supp_content}

\end{document}

%% file: supp_content.tex
\section{Results for Additional Losses}\label{sec:appendix_additional_losses}

In Table \ref{tab:new_losses} we present results for two additional losses which are described below.
Both losses are meant for the ``single positive only" setting. 

\noindent
\textbf{Pairwise ranking.} 
The pairwise ranking (PR) loss is minimized when all of the predictions for positive labels are larger than all of the predictions for negative labels  \cite{gong2014deep}. 
However, in the ``single positive only" setting we have many unobserved labels which may be either positive or negative.
If we assume that all unobserved labels are negative, then the PR loss becomes 
\begin{align*}
    \mathcal{L}_\mathrm{PR}(\mathbf{f}_n, \mathbf{z}_n) &= \sum_{i,j = 1}^L \indic{z_{ni} = 1, z_{nj} = 0} \max(0, 1 - f_{ni} + f_{nj}).
\end{align*}
The results in Table \ref{tab:new_losses} indicate that this straightforward adaption of $\mathcal{L}_\mathrm{PR}$ to the ``single positive only" setting does not outperform the $\mathcal{L}_\mathrm{AN}$ baseline. Different loss weighting schemes or label smoothing may be beneficial for $\mathcal{L}_\mathrm{PR}$, but studying the effect of those modifications is beyond the scope of our work.

\noindent
\textbf{Assume negative with asymmetric label smoothing.} 
In the binary classification setting, traditional label smoothing replaces the ``hard" target distributions with ``smoothed" target distributions according to the mapping
\begin{align*}
    (1, 0) &\to (1-\epsilon/2, \epsilon/2)\\
    (0, 1) &\to (\epsilon/2, 1-\epsilon/2)
\end{align*}
where $\epsilon \in [0, 1]$ is a hyperparameter. The hyperparameter $\epsilon$ controls the training targets for both positive and negative labels. However, it may not always make sense to treat positive labels and negative labels the same way. To study this, we modify traditional label smoothing so that it modulates positive and negative labels independently, replacing the single hyperparameter $\epsilon$ with with two hyperparameters $\epsilon_p$ (for positive labels) and $\epsilon_n$ (for negative labels).
This asymmetric label smoothing loss is given by 
\begin{align*}
    \mathcal{L}_{\mathrm{AN-LS}(\epsilon_p, \epsilon_n)}(\mathbf{f}_n, \mathbf{z}_n) = -\frac{1}{L} \sum_{i=1}^L[ & \indic{z_{ni}=1}^{\frac{\epsilon_p}{2}, \frac{\epsilon_n}{2}} \log(f_{ni})  \\
    + & \indic{z_{ni} \neq 1}^{\frac{\epsilon_n}{2}, \frac{\epsilon_p}{2}} \log(1-f_{ni}) ]
\end{align*}
where $\epsilon_p, \epsilon_n \in (0, 1)$ are hyperparameters and 
\[ \indic{Q}^{\alpha, \beta} = (1-\alpha) \indic{Q} + \beta \indic{\neg Q}. \] 
For instance, the loss $\mathcal{L}_{\mathrm{AN-LS}(0, 0.1)}$ applies label smoothing (with standard smoothing parameter value 0.1) to the negative labels but leaves the positive labels untouched. 
This is a reasonable approach to the ``assume negative" case because the positive labels are assumed to be reliable while the negative labels may be incorrect. 
Table \ref{tab:new_losses} shows that $\mathcal{L}_{\mathrm{AN-LS}(0, 0.1)}$ is better than $\mathcal{L}_\mathrm{AN-LS}$ in the fine-tuned case (+1.0 MAP for VOC12, +0.7 MAP for COCO) and worse in the linear case (-0.3 MAP for VOC12, -0.4 MAP for COCO). 
At least in the fine-tuned case, asymmetric label smoothing seems to effectively counteract the special kind of label noise we encounter in the ``single positive only" setting. 

\begin{table}
\centering
\small
\begin{tabular}{|l| c c | c c |}
\hline
& \multicolumn{2}{|c|}{Linear} & \multicolumn{2}{|c|}{Fine-Tuned} \\ \hline 
Loss & VOC12 & COCO & VOC12 & COCO \\ \hline
$\mathcal{L}_{\mathrm{AN}}$ & 84.2 & 62.3 & 85.1 & 64.1\\
$\mathcal{L}_\mathrm{PR}$ & 84.2 & 56.9 & 81.5 & 56.3\\
$\mathcal{L}_{\mathrm{AN-LS}}$ & 85.3 & 64.8 & 86.7 & 66.9\\
$\mathcal{L}_{\mathrm{AN-LS}(0, 0.1)}$ & 85.0 & 64.4 & 87.7 & 67.6 \\
\hline
\end{tabular}
\vspace{5pt}
\caption{
Results on VOC12 and COCO for two additional losses: pairwise ranking ($\mathcal{L}_\mathrm{PR}$) and asymmetric label smoothing ($\mathcal{L}_\mathrm{AN-LS}(\epsilon_p, \epsilon_n)$). The losses are described in Section \ref{sec:appendix_additional_losses}. For ease of reference, we also include results for $\mathcal{L}_\mathrm{AN}$ and $\mathcal{L}_\mathrm{AN-LS}$ from Table \ref{tab:natimg_compare} in the main paper. Note that all results are for the ``single positive only" setting.
}
\label{tab:new_losses}
\end{table}

\section{Effect of Misspecified $k$}\label{sec:appendix_misspecified_k}

In the main paper, losses that utilize the parameter $k$ are provided with the ``true" value $k^*$ computed over the fully labeled training set. 
In practical settings it may be necessary to estimate $k$ based on domain knowledge or a small sample of fully labeled images resulting in a value of $k$ which deviates from $k^*$.
In this section we consider the effect of these deviations.

\begin{figure}
    \centering
    \includegraphics[width=0.45\textwidth]{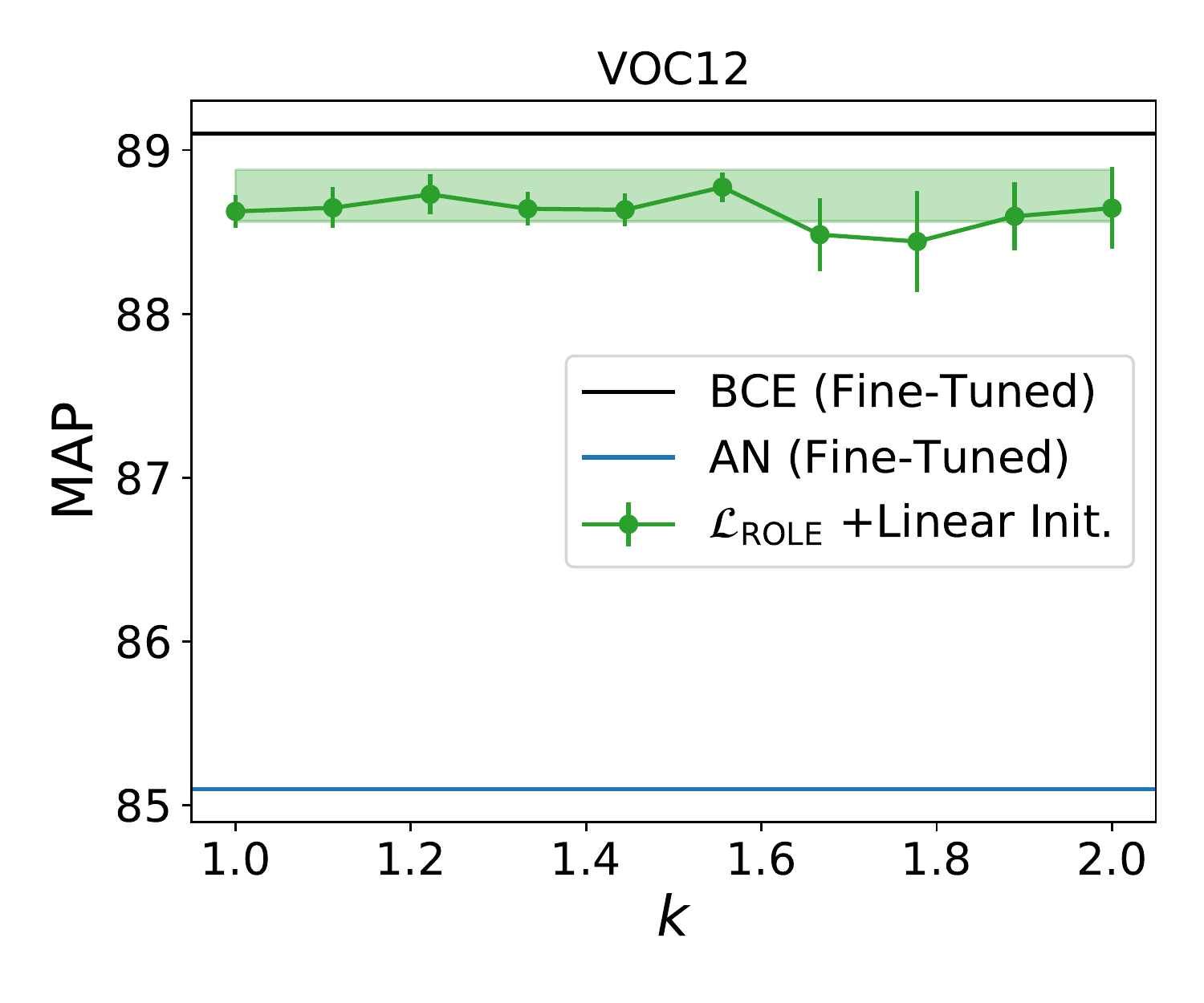} \\
    \includegraphics[width=0.45\textwidth]{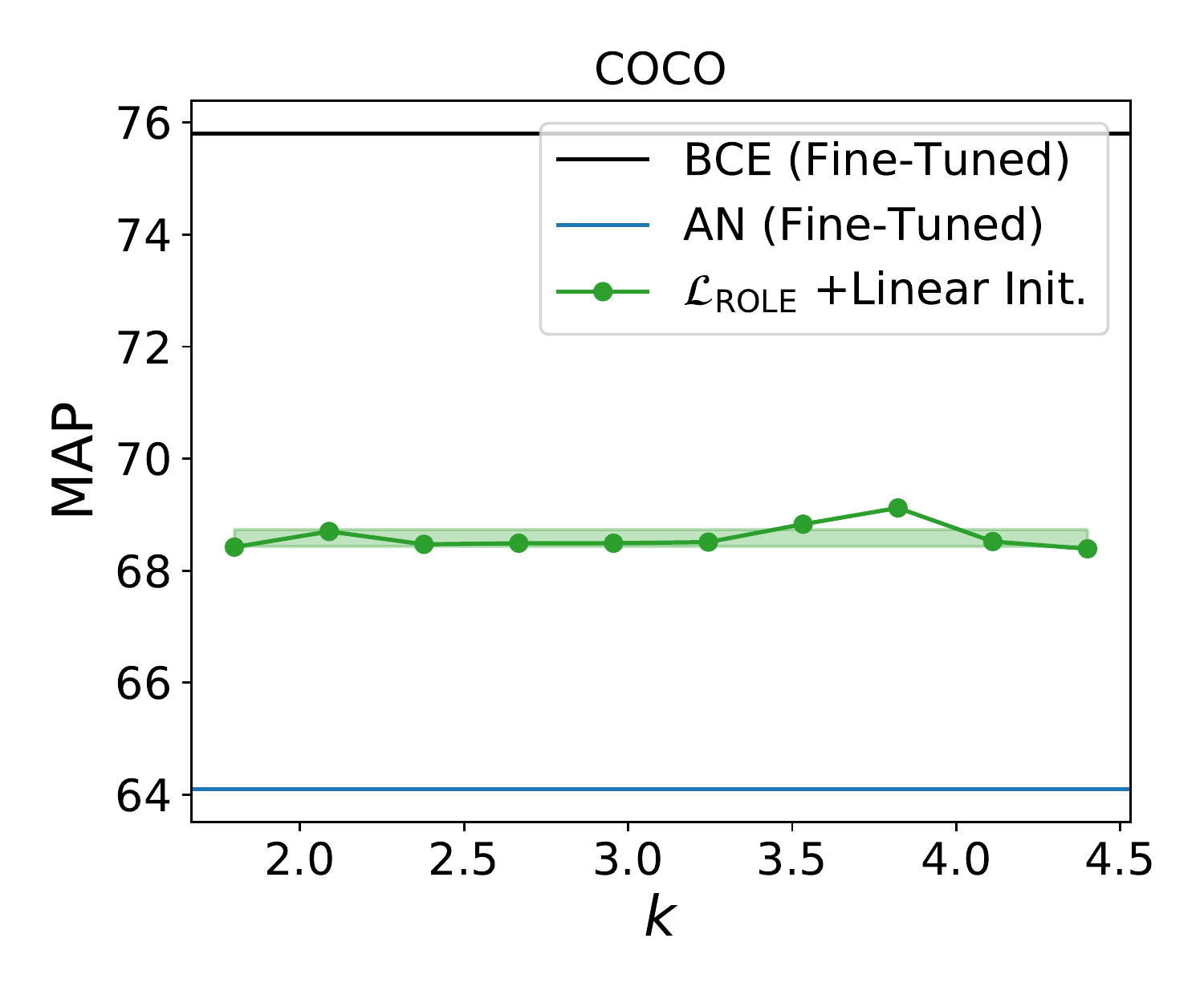} \\
    \caption{
    Performance of $\mathcal{L}_\mathrm{ROLE}$ as a function of $k$.
    We consider 10 linearly spaced $k$ values in the interval $I_5$ from Table \ref{tab:k_estimates}. 
    For VOC12 (top) we show mean and standard deviation over 5 runs at each value of $k$, while for COCO (bottom) we show only one run per value of $k$. 
    The shaded band indicates the performance of $\mathcal{L}_\mathrm{ROLE}$ +LinearInit with the true value of $k$ (mean $\pm$ std. computed over 5 runs). 
    In the main paper we use $k=1.5$ for VOC12 and $k=2.9$ for COCO (as shown in Table \ref{tab:data_stats}).
    }
    \label{fig:misspecified_k}
\end{figure}

Suppose we are interested in estimating $k$ from data. 
Given $M$ fully labeled examples $S_M = \{(\mathbf{x}_n, \mathbf{y}_n)\}_{n=1}^M$, we can empirically estimate $k$ as 
\begin{align}
    \hat{k}_M = \frac{1}{M} \sum_{n = 1}^M \sum_{i=1}^L y_{ni}. \label{eq:k_hat_M}
\end{align}
To understand whether $k$ can be estimated well from $M$ fully labeled examples, we need to know something about the distribution of $\hat{k}_M$. 
We first generate $T = 10^5$ realizations of $S_M$, denoted $S_M^{(1)}, \ldots, S_M^{(T)}$.
Each realization of $S_M$ consists of $M$ random examples from the training set drawn uniformly at random (without replacement). 
We then compute the collection of estimates $\mathcal{K}_M = \{\hat{k}_M^{(1)}, \ldots, \hat{k}_M^{(T)}\}$ where $\hat{k}_M^{(t)}$ is computed from $S_M^{(t)}$ using Equation \ref{eq:k_hat_M} for $t = 1,\ldots, T$. 
Finally, we take the 5th and 95th percentiles of $\mathcal{K}_M$ to produce an approximate 95\% confidence interval $I_M$ for $\hat{k}_M$. 
In Table \ref{tab:k_estimates} we provide $I_M$ for $M\in \{5, 10, 25\}$. 
In Figure \ref{fig:misspecified_k} we show the performance for $\mathcal{L}_\mathrm{ROLE}$ +LinearInit. over $I_5$. 
We can see that performance is stable for values of $k$ in $I_5$. 
Put another way, the performance of $\mathcal{L}_\mathrm{ROLE}$ +LinearInit. is likely to be similar whether we use $k^*$ or an estimate $\hat{k}_5$ based on only five fully labeled training examples.  

\begin{table}[h]
\centering
\small
\begin{tabular}{|l|c c c|}\hline
     & \multicolumn{3}{c|}{95\% CIs for $\hat{k}_M$} \\
     Dataset & $I_{5}$ & $I_{10}$ & $I_{25}$ \\ \hline
     VOC12 & (1.0, 2.0) & (1.1, 1.9) & (1.2, 1.7) \\
     COCO & (1.8, 4.4) & (2.1, 4.0) & (2.4, 3.6) \\
     NUS & (0.8, 3.2) & (1.0, 2.8) & (1.3, 2.5) \\
     CUB & (25.2, 37.8) & (27.0, 35.9) & (28.6, 34.2) \\ \hline
\end{tabular}
\vspace{5pt}
\caption{
95\% confidence intervals $I_M$ for estimates $\hat{k}_M$ of $k$ based on $M \in \{5, 10, 25\}$ fully labeled training examples. 
We take these intervals to represent ranges of ``reasonable" values for $k$ - see Section \ref{sec:appendix_misspecified_k} for details.
All confidence intervals are computed based on $10^5$ trials.
}
\label{tab:k_estimates}
\end{table}

\section{Dataset Statistics}
In Table \ref{tab:data_stats} we give some basic statistics for our multilabel image classification datasets. 
The average number of positive labels per image is computed on the (fully labeled) training set - this is the ``true" $k$ (referred to as $k^*$ in Section \ref{sec:appendix_misspecified_k}) we use for $\mathcal{L}_\mathrm{ROLE}$ and $\mathcal{L}_\mathrm{EPR}$ in the main paper. 
We study the effect of setting $k$ to other values in Section \ref{sec:appendix_misspecified_k}.

\begin{table}[h]
\centering
\small
\begin{tabular}{|l|c|c c c|c c c|}\hline
    & & \multicolumn{3}{c|}{\# Pos. Labels / Image} \\
     Dataset & \# Classes & Max. & Min. & Avg. \\ \hline
     VOC12 & 20 & 5 & 1 & 1.5 \\
     COCO & 80 & 18 & 1 & 2.9 \\
     NUS & 81 & 11 & 0 & 1.9 \\
     CUB & 312 & 71 & 3 & 31.4 \\ \hline
\end{tabular}
\vspace{5pt}
\caption{
Label statistics for each dataset, including the maximum, minimum, and average ($k$) number of positives per image.
}
\label{tab:data_stats}
\end{table}